\documentclass[11pt]{article}
\usepackage{acl2012}
\usepackage{times}
\usepackage{latexsym}
\usepackage{amsmath}
\usepackage{multirow}
\usepackage{url}
\usepackage{latexsym}
\usepackage{graphicx}
\usepackage{algorithmic}
\newcommand{\comment}[1]{}

\newcommand\T{\rule{0pt}{2.2ex}}
\newcommand\B{\rule[-1.0ex]{0pt}{0pt}}
\title{Tiered Clustering to Improve Lexical Entailment}

\author{John Wieting \\
  University of Illinois-Urbana Champaign \\
  {\tt wieting2@illinois.edu}}

\date{}

\begin{document}
\maketitle
\begin{abstract}
Many tasks in Natural Language Processing involve recognizing lexical entailment. Two different approaches to this problem have been proposed recently that are quite different from each other. The first is an asymmetric similarity measure designed to give high scores when the contexts of the narrower term in the entailment are a subset of those of the broader term. The second is a supervised approach where a classifier is learned to predict entailment given a concatenated latent vector representation of the word. Both of these approaches are vector space models that use a single context vector as a representation of the word. In this work, I study the effects of clustering words into senses and using these multiple context vectors to infer entailment using extensions of these two algorithms. I find that this approach offers some improvement to these entailment algorithms.

\end{abstract}

\section{Introduction}

An important task in Natural Language Processing research is Recognizing Textual Entailment (RTE). This is because this task is very relevant for the problems of text summarization, information retrieval, information extraction, question answering, and many others. An RTE problem involves a pair of sentences, the first of which is known as the text and the second is known as the hypothesis. The goal of the task is to determine whether the text entails the hypothesis, or in other words to determine whether the meaning of the hypothesis can be inferred from the text. An example, from \cite{TurneySa13} would be:

$$
    \text{T: George was bitten by a dog.}
$$
$$
    \text{H: George was attacked by an animal}    
$$

Clearly in this example, H can be inferred from T. However, notice that the converse is not true as we would have no way to know that the manner in which George was attacked was by being bitten or that the animal who attacked him was a dog. Thus the RTE task involves an asymmetric relation between sentences.

It also important to note that in order for a system to obtain the correct answer for this problem it would likely have to determine that {\it bitten} entails {\it attacked} and { \it dog} entails {\it animal}. Thus entailment must function well at the word level. This task is known as lexical entailment.

Recently, in \cite{TurneySa13} three different approaches to lexical entailment were analyzed. Two of them had been recently proposed and the third is one of the contributions of that paper. The first, known as balAPinc (balanced average precision for distributional inclusion) \cite{Kotlerman10}, is an asymmetric similarity measure and the second, ConVecs (concatenated vectors) \cite{BaroniBDS12} uses a supervised approach. Both of these algorithms are vector space models in that they both rely on a context vector representation of the words as an input. 

Recent progress has been made in word similarity, another task that uses vector space models, by clustering together word senses and using these clusters to determine their similarity score \cite{Reisinger10na} and \cite{Reisinger10em}. The reason for separating out these senses is that many words are homonymous (contain multiple unrelated meaning) or polysemous (contain multiple related meanings). By representing all of these meanings in one single vector we are creating a noisy signal for that word and the signal may perform badly when one of the lesser used meanings of the word is to be scored.

Naturally, this idea should also carry over to lexical entailment, which to the best of my knowledge has not yet happened. Thus in this paper, I modify the lexical entailment algorithms above to take into account word sense clusters. I also investigate two approaches in order to accomplish the clustering. The first follows that in \cite{Reisinger10em}, in a technique known as tiered clustering. This is a Dirichlet Process clustering model that is equivalent to the nested Chinese Restaurant Proccess \cite{Blei04} with a fixed depth of two. One potential weakness of this model is that it only uses word counts to form clusters. Thus I use a variation of correlaton clustering \cite{Bansal02} which is useful for cases when one wants to cluster solely using a distance metric. An alternative could have been k-medioids, however the advantage of correlation clustering is that it is much faster which is very important as clustering must be done for each word in the vocabulary. Additionally, the number of clusters does not need to be specified a priori with correlation clustering.

Using these models, improvement was made in these state of the art entailment algorithms. The improvement, not surprisingly, is dependent on how the word senses are used to make a classification decision. Interestingly, just choosing the maximum score over all word senses is not the best approach and instead some type of averaging over the scores or representations tends to give better results.

The remainder of this paper is organized as follows: Section 2 provides background information on the entailment and clustering algorithms used in this paper, Section 3 illustrates the extensions that were done to the entailment algorithms to incorporate the word sense clusters into the classification decision, Section 4 details the experimental setup, Section 5 discusses the results, and Section 6 concludes.

\section{Background}

\subsection{Defining Lexical Entailment}
Given two sentences, whether or not there exists an entailment relation between them is more of a matter of common sense than logic. Thus it is not easy to define lexical entailment in a useful way. \cite{Zhitomirsky-GeffetD09} defined entailment in terms of substitution. Essentially they say word {\it a} entails word {\it b} if {\it a} can substitute for {\it b} in a sentence and this new sentence entails the original. This approach leads to high inter-annotater aggreement in the entailment task, however it is argued that this definition does not cover all cases of entailment. For instance, Turney et. al. argue that in the sentences {\it Jane dropped the glass} and {\it Jane dropped something fragile}, the word {\it glass} should entail {\it fragile}. They then go on to define entailment through the semantic relations in \cite{Bejar91}. They claim that some of these relations define an entailment relationship between all pairs of words having that relation. Thus if one solves semantic relation identification, they also solve lexical entailment. These two definitions have motivated algorithms as well as data sets for the entailment task. In this paper, we investigate an algorithm motivated by the first definition (balAPincs) and one by the second (ConVecs). Also we evaluate these models on data sets that were also motivated by these definitions. The first, known as BBDS \cite{BaroniBDS12} is motivated by the first definition and is largely a collection of hyponym-hypernym pairs. The second, known as JMTH \cite{TurneySa13} is motivated by the second definition and is a very difficult data set due to the more expressive definition of entailment.

\subsection{Approaches for Lexical Entailment}
    
\subsubsection{balAPinc}
This approach, first described in \cite{Kotlerman10}, aims to reward those situations when the first context vector argument is a subset of the second. In other words, the features of the first context vectors should be included in the second. This idea naturally comes from the distributional inclusion hypothesis \cite{Geffet05}, which states that if word {\it a} occurs in a subset of the context of word {\it b} then {\it a} often entails {\it b}. The formula for calculating balAPinc is below. $F_i$ denotes the context vector where all nonzero entries have been removed. In practice, this feature vector includes only the top 1000 or so features to prevent lots of low occurring features to influence the score.

\begin{equation}
\text{balAPinc}(u,v)=\sqrt{\text{APinc}(u,v)\cdot\text{LIN}(u,v)}
\end{equation}

\begin{equation}
\text{APinc}(u,v)=\frac{\sum_{r=1}^{\vert F_u \vert}{\vert P(r, F_u, F_v)\cdot rel(f_{ur}, F_v) \vert}}{\vert F_u \vert}
\end{equation}

\begin{equation}
\text{P}(r,F_u,F_v) = \frac{\vert \text{inc}(r,F_u,F_v) \vert}{r}
\end{equation}

\begin{equation}
\text{rel}(f,F_w) = \begin{cases}
1-\frac{\text{rank}(f,F_w)}{\vert F_w \vert + 1} &\text{if $f \in F_w$}\\
0 &\text{otherwise}
\end{cases}
\end{equation}

\begin{equation}
\text{inc}(r,F_u,F_v) = \{ f \vert \text{rank}(f,F_u) \leq r \text{, } f \in (F_u \cap F_v) \}
\end{equation}

\subsubsection{ConVecs}
ConVecs \cite{BaroniBDS12} operates under the hypothsis that the entailment of words {\it a} and {\it b} is a learnable function of the concatenation of their context vectors. The authors propose using Singular Value Decomposition to reduce the dimensionality of the context vectors and then use a kernal SVM to learn this function. In order to obtain a score from this model, probability estimate can be used and the probability for the positive class can be used as the score.

\subsection{Clustering Occurrences}
The idea of clustering word senses has been used before in the task of determining word similarity. In \cite{Reisinger10na}, the authors use the mixture of von Mises-Fisher distributions (movMF) clustering method \cite{Banerjee05}. One disadvantage of this approach is that the number of clusters must be specified in advance. To remove this restriction, \cite{Reisinger10em} introduces tiered clustering, a nonparametric model, that does not require this parameter to be specified. In this paper, I also explore the effectiveness of another type of clustering algorithm, correlation clustering, which also has the advantage of not needing this parameter to be specified. The advantage this algorithm would have over tiered clustering is that it takes into account the relatedness of the words themselves when clustering and not just their counts.

\subsubsection{Tiered Clustering}

\begin{figure}[!ht]
\includegraphics[height=3cm, width=7cm]{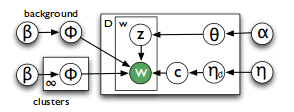}
\caption{
Plate diagram for tiered clustering \cite{Reisinger10em}.
}
\end{figure}

The plate diagram for this Bayesian clustering model is in Figure 1 above. Inference for tiered clustering is done using a collapsed Gibbs sampler. As this model is a special case of the nested Chinese Restaurant Process \cite{Blei04} where the depth is fixed at 2. There are two advantages this approach has over other clustering methods. The first, which has already been mentioned, is that there is no need to specify a priori the number of clusters. The second is that there is a root node which can capture those features which have context-independent variance, preventing these features from dominating the clustering process.

The sampling algorithm is similar to that of the NCRP. Essentially, in one iteration of the Gibbs sampling method, we first sample a cluster for the current word occurrence (or document in the terminology of NCRP). This sampling could result in the creation of a new cluster. Then for each feature (or word in the terminology of NCRP) we sample the node to which it belong. There are only two choices here since the path is only two nodes long. Either it lies in the root or in the cluster to which its document belongs. The structure with the lowest log-likelihood when the maximum number of iterations is reached is the output of the algorithm.

%\begin{equation}

%\end{equation}

%\begin{equation}
%\begin{split}
%P(z_{i,d}=t \vert {\bf z_{-(i,d)}}, {\bf w}, \alpha, \beta) &= \\ \frac{(\eta_t^{(w_{i,d})} + \beta) \cdot (\eta_t^(d)+\alpha)}{(\sum_w \eta_t^{w} + \beta) \cdot (\sum_j \eta)j^{(d)} + \alpha)}
%\end{split}
%\end{equation}

\subsubsection{Correlation Clustering}
In order to incorporate the relatedness of words into the clustering. I used a variation of correlation clustering \cite{Bansal02}. The algorithm is very straightforward and all that is needed is a single parameter $\sigma$, and a similarity metric. Basically a single point is drawn from the set of points that have not yet been assigned a cluster. Then every other point is compared to this one using the similarity metric, and if the score of this pair is greater than $\sigma$ then these points are placed into the same cluster. The process is repeated until all points have been assigned a cluster. However due to the large amount of time this can take if there are numerous outlier points and also to limit the number of clusters, I added a termination condition. The algorithm would terminate after it had at least two clusters each containing at least 2.5\% of the points and if the last five clusters that had been formed contained less than 2.5\% of the points. The idea here is that the algorithm will likely generate the largest clusters first and then when most of the remaining points are outliers, it will terminate as it will be unable to create any other large clusters. This saves a lot of computation time. The disadvantage of this algorithm is that it is a greedy algorithm and it could miss out on some nice clusters. Also interesting, the same point can contribute to multiple clusters.

The distance function used was the LLM measure used in \cite{DRSTV09}. The equation is below where $S_2$ refers to the smaller of the two context vectors. To compute the distance between words, the Wu Palmer algorithm was used \cite{Wu94}. This metric returns a score for the words based on their similarity in WordNet.

\begin{equation}
\text{LLM}(S_1,S_2)= \frac{\sum_{v \in S_2} \text{max}_{u \in S_1} \text{sim}(u,v)}{\vert S_2 \vert}
\end{equation}

\section{Algorithm Extensions}

Both balApincs and ConVecs were designed to accomodate a single context vector representative for each word. In order to accomodate the multiple senses a word can have the algorithms must be extended for this more general case.

balAPincs can be extended in two obvious ways. In the first way, we can compute the score of all possible pairs of word senses between the two words and use the maximum score for these words. Support for this idea lies in \cite{TurneySa13} where the author claims that two words entail each other if any pair of their senses entail each other. It is also mentioned in \cite{Reisinger10na}. Another idea is to use the average of these scores. This approach is used in \cite{Reisinger10em} and \cite{Reisinger10na}. Despite recombining the senses, this approach tends to work well because it is more robust to noise. Reisinger argures that this approach is still better than using single prototypes because now less often used senses have more influence. Both of these approaches can also be extended by weighting the clusters by their probability or in other words incorporating a prior based on how many word occurances are in that cluster compared to all word occurances encountered. These extensions are mentioned and used in \cite{Reisinger10na} but are not used in the tiered clustering paper \cite{Reisinger10em}. One issue with this weighting is that it would seem that there would be little difference between this approach and not clustering at all. In this paper, both the averaging and maximum approaches are explored. 

Convecs can be extended as well in a number of ways to account for multiple context vectors per word. One issue for ConVecs is that since it is a supervised method, how do we define positive and negative examples when dealing with word senses? There are several ways this can be done as well. The first is to find the word sense pairs for each word that overlap the most using balAPincs. Then if the example is a positive example, this pair would represent that example as is it is most likely to exhibit the entailment relation. Similarly, if the example is supposed to be negative, this pair should also represent the example as we want our negative examples to lie as close to the margin as possible in order to learn a model with good generalization properties. Another approach would be to simply average the vectors of all word senses and use that as the example. For evaluation, again we have several choices. As in balAPinc, we could average the scores or choose the maximum score for each example. Another approach would be to average the feature vectors and then use the result of applying the classifier to this vector. All three of these approaches are explored in this paper.

\section{Experiments}

For evaluation, 10 fold cross validation was performed on two of the three data sets used in \cite{TurneySa13}. These data sets were chosen because they both were created with different definintions of lexical entailment in mind, giving us an opportunity to evaluate this approach in light of two different philosophies on entailment. The first, dubbed BBDS and was also used in \cite{BaroniBDS12}, consists of 1228 examples. This dataset was balanced exactly to contain equal numbers of positive and negative examples and was created using the substitution definition of entailment.  The second data set with 720 examples, JMTH, was created from the SemEval-2012 Task 2 following the instruction in \cite{TurneySa13}. This is a difficult data set containing such positive examples as {\it crack} entails {\it glass}. It was created using the semantic relation definition of entailment.

Data for the experiment in the form of tagged frames around word occurrences was used in \cite{TurneyP10} and was given to us by request from the author. A window size of four on both sides of the occurrence was used. The context matrices created from this data were created in the same fashion as the one used in that paper. Each row corresponds to a term and the columns represent the context of the word occurrences in the form of unigrams. There are 139,246 columns, each a unigram indicating if that context had appeared to the left or right of the target word in the occurrence. The context matrix in that paper also had 114,501 terms which is far too expensive to compute when we are also taking word senses into account. Thus the 2,385 terms included in the evaluation data sets were used to create the matrices.

There were at most 10,000 occurrences for each term. Out of these 10,000 (or less) occurrences, 1000 were sampled to create the context matrices. These were chosen by taking those sentences which contained the most context words as these would provide more interesting and informative clusters. An effort was made to pick unique sentences as after initial experiments it became clear that some sentences were included in these occurrences multiple (sometimes more than a thousand) times. This pruning of sentences was done so that the clustering algorithms would have less data to cluster and would not take as long. It also eliminated one sentence being repeated many times and influencing the context vectors disproportionally.

After 1000 occurrences had been chosen, the left and right contexts for each word were merged in order to reduce sparsity. Additionally, these features were also pruned as per \cite{Reisinger10em} to only the most frequent 500 terms. This was deemed sufficient as the only features allowed were those that were columns in our matrix. Hence stop words and other high frequency artifacts were removed. 

Correlation clustering was accomplished using a $\sigma$ value of 0.85. This parameter was lightly tuned until it produced attractive clusters on a few homonyms. The parameters used in Tiered clustering were $\alpha$=1.0, $\beta$=0.1, and $\eta$=0.01 in an attempt to keep the number of clusters per word to a minimum. Gibbs sampling was done for 12,000 iterations for each word.

After clustering, only those clusters which contained at least 2.5\% of the occurrences were kept. The instances in the vectors were then mapped to their original vectors so all features would be present for classification. Then all the occurrences in the clusters were combined by adding together their context vectors, and this combination was a prototype for a particular sense of the word. These prototypes were used to create the context matrix. It is important to note that the context matrices that resulted did not just contain counts. The oc-occurrence frequencies were transformed to positive pointwise mutual information values (PPMI) \cite{TurneyP10} in order to better represent the importance of a context feature for a given word.

balAPincs was trained by picking the optimal threshold on the training data. Both average and maximum approaches were used in order to attempt to determine which was better. ConVecs used 100 latent features for each word and was trained with a quadratic kernel using LibSVM. Both the average and maximum  were used in evaluation as well where the positive and negative examples were determined by balAPincs. Additionally a new approach where the feature vectors were averaged was also done with ConVecs giving the best results.

In order to check whether clustering these word occurrences improved performance, a baseline approach was used where a single prototype for each word was constructed from the 1000 occurrences. The results of the experiments are shown in Tables 1 and 2 below. Accuracy was used to compare the different approaches because the data sets were completely balanced.

\begin{table*}[!ht]
\begin{small}
\centering
\begin{tabular} { | l || l | l ||} \hline
\T \B {\bf Cluster} &\multicolumn{2}{|c|}{\bf Accuracy}\\
\hline\hline
\T & BalAPincs & Convecs \\\hline
\bf Baseline  & 68.1 & 75.6 \\
\bf Correlation Clusters AvgScore & 67.3 &  71.5\\
\bf Correlation Clusters MaxScore & 66.5 & 66.7\\
\bf Correlation Clusters AvgVector & NA & 74.8\\ 
\bf Tiered Clusters AvgScore & \bf 68.2 & 71.5\\
\bf Tiered Clusters MaxScore & 65.0 & 51.2\\
\bf Tiered Clusters AvgVector & NA & \bf 76.4\\
\hline
\end{tabular}
\caption{
Comparison of Algorithms on BBDS
}
\end{small} % end small
\end{table*}

\begin{table*}[!ht]
\begin{small}
\centering
\begin{tabular} { | l || l | l ||} \hline
\T \B {\bf Cluster} &\multicolumn{2}{|c|}{\bf Accuracy}\\
\hline\hline
\T & BalAPincs & Convecs \\\hline
\bf Baseline  & 55.7 & 61.1 \\
\bf Correlation Clusters AvgScore & 51.8 & 61.1\\
\bf Correlation Clusters MaxScore & 55.8 & 50.0 \\
\bf Correlation Clusters AvgVector & NA & \bf 68.1 \\ 
\bf Tiered Clusters AvgScore & 56.8 & 62.5 \\
\bf Tiered Clusters MaxScore & \bf 57.6 & 50.0 \\
\bf Tiered Clusters AvgVector & NA & 66.7 \\
\hline
\end{tabular}
\caption{
Comparison of Algorithms on JMTH
}
\end{small} % end small
\end{table*}
\section{Discussion}

\begin{figure}[!ht]
\includegraphics[width=90mm]{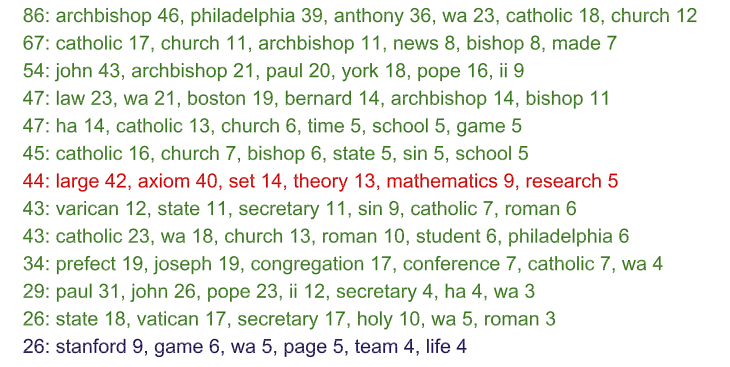}
\caption{
Clusters for the word {\it cardinal} from a modified correlation clustering algorithm. The first number is the number of documents (occurrences) in that cluster.
}
\end{figure}

\begin{figure}[!ht]
\includegraphics[width=90mm]{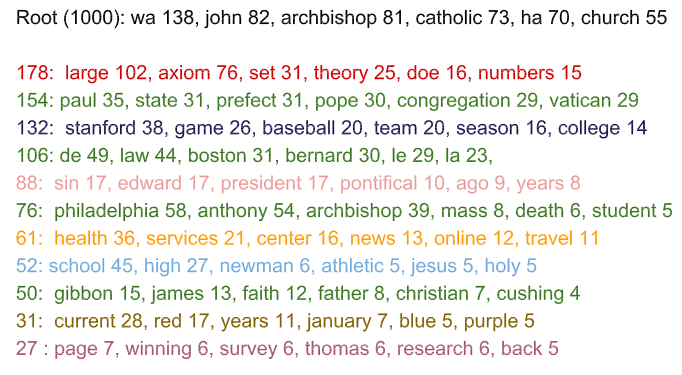}
\caption{
Clusters for the word {\it cardinal} from the tiered clustering algorithm. The {\it Root} refers the the root node in the model that is meant to capture the background features in the occurrences. Notice how it contains words like {\it wa} which are likely artifacts from tokenization.
}
\end{figure}

\subsection{Results}
The results from the experiments are mixed. One thing that is clear from the results is the importance of how the senses are combined in order to make a classification decision. Suprisiningly, just taking the maximum score is not always the best option. It gives very inconsistent results that likely has to do with the clustering as well as the noise in the data. This is best illustrated with ConVecs as this approach was able to find a cluster that gave a positive signal in every example in the data set, hence the 50.0\% accuracy. Thus some averaging tends to reduce the effects of this noise while giving senses that appear less often in the data to have more influence and affect the decision more than they would in a single prototype approach. 

From the experiments, though it seems that clustering does do better than the baseline if the appropriate algorithm is chosen. For instance, with ConVecs, tiered clustering with averaging the vectors beats the baseline in both data sets. Similarly, with balAPinc, tiered clustering using the average score is also better than the baseline as well, although not significantly for the BBDS data set. It is also interesting to see that clustering has a bigger effect on performance on the more difficult data set, JMTH.

I believe that this type of clustering is useful, even though it comes with a large memory and performance hit. However, the need to average to achieve good results means more work would need to be done if applying these techniques on a real-world problem where there is some context available to deduce the sense of the word with some accuracy. I think this would be an interesting problem to explore as just taking a single cluster would likely not give the best performance and perhaps a distribution of clusters would need to be used to make the scoring decision.

\subsection{Clusters}
Clusters from correlation clustering and tiered clustering are shown above in Figures 2 and 3 repectively. These figures are interesting for a couple of reasons. First of all, I was initially skeptical of the root node in the tiered clustering algorithm as I figured that the features were already heavily filtered prior to clustering and hence the root node would only serve to eliminate useful features that can be used to assign the occurrences to clusters. However, the root does seem to capture noisy features like those that appear to be artifacts of tokenization. Granted, these features could probably be pruned out using some kind of tf-idf, but it seems that this node does have a useful function. 

Another thing to point out is that there are larger differences between clusters in the tiered clustering approach versus the correlation clustering approach. Hence it appears tiered clustering is producing better clusters overall. However, tiered clustering produces an average of 15.6 clusters per occurrence while correlated clustering does 7.5. Furthermore, tiered clustering took an average of 7.5 minutes to cluster while correlated clustering took an average of 1.5 minutes. This is important because the memory used in this model scales linearly with the average number of clusters per word and the compexity of evaluating entailment has a quadratic relationship with the average number of clusters.

Lastly, as a reflection of the word occurrence quality, notice that an important sense of {\it cardinal} as a bird is missing. This illustrates an important point about context vectors in that the distribution of contexts is based on what exists in the data (in this case, the crawling was done on university web pages). I think there is room for improvment by perhaps crawling wikipedia instead which may give a more natural distribution of word occurrences.

\section{Conclusion}

This paper applied two clustering techniques for clustering word senses in an effort to improve two state of the art lexical entailment techniques. The results showed that clustering word senses does provide some improvement, but care must be taken in combining the senses in order to make a classification decision. The experiments conducted in this paper show that tiered clustering, with an appropriate algorithm for combining sense, consistantly can give better results over the single prototype baseline. 

There is a lot of room here for future work in this arena. For one, the clustering could still be improved. It would be interesting to see if there was a way to incorporate word relatedness into the tiered clustering model. By merging some of the similar clusters, this approach would be more resilient as the number of clusters would be more limited. Another way to limit the clusters would be to have a higher threshold of the amount of occurrences needed for a cluster to be kept. Perhaps tuning this parameter would have produced better results. There are other avenues for improvement as well such as using less noisy and more representative data, and finding better ways of use these clusters to produce a score. The latter is especially deserving of further study. In conclusion, I think that this is a viable approach to improving lexical entailment, but further work must be done for the benefits of this approach to be worth the large computational costs.

\section*{Acknowledgments}

Thanks to Prof. Hockenmaier, Joseph Reisinger, and Prof. Roth for their helpful discussions. Also thanks to Dr. Turney and Prof. Baroni for providing the data used in these experiments.

\bibliography{ref}

\bibliographystyle{acl2012}

\end{document}